# Multi-task Joint Strategies of Self-supervised Representation Learning on Biomedical Networks for Drug Discovery

Xiaoqi Wang[1†], Yingjie Cheng[1†], Yaning Yang[1], Yue Yu[2], Fei Li[3*], and Shaoliang Peng[1,2,4*]

[1]College of Computer Science and Electronic Engineering, Hunan University, Changsha 410082, China.
[2]Institute of Open Source Software and Platform, Peng Cheng Laboratory, Shenzhen, 518000, China.
[3]Computer Network Information Center, Chinese Academy of Sciences, Beijing 100850, China.
[4]The State Key Laboratory of Chemo/Biosensing and Chemometrics, Hunan University, Changsha 410082, China.

[†]These authors contributed equally to this work.
[*]To whom correspondence should be addressed; E-mail:
    pittacus@gmail.com (Fei Li);
    and slpeng@hnu.edu.cn (Shaoliang Peng).

## Abstract

Self-supervised representation learning (SSL) on biomedical networks provides new opportunities for drug discovery. However, how to effectively combine multiple SSL models is still challenging and has been rarely explored. Therefore, we propose multi-task joint strategies of self-supervised representation learning on biomedical networks for drug discovery, named MSSL2drug. We design six basic SSL tasks inspired by various modality features including structures, semantics, and attributes in heterogeneous biomedical networks. Importantly, fifteen combinations of multiple tasks are evaluated by a graph attention-based multi-task adversarial learning framework in two drug discovery scenarios. The results suggest two important findings. (1) Combinations of multimodal tasks achieve the best performance compared to other multi-task joint models. (2) The local-global combination models yield higher performance than random two-task combinations when there are the same size of modalities. Therefore, we conjecture that the multimodal and local-global combination strategies can be treated as the guideline of multi-task SSL for drug discovery.

## 1 Introduction

Drug discovery is an important task for improving the quality of human life. However, it is an expensive, time-consuming, and complicated process that has a high chance of failure [1-2]. To improve the efficiency of drug discovery, a great number of researchers have devoted to developing or leveraging deep learning to speed up its intermediate steps, such as molecular property predictions [3-4], drug-target interaction (DTI) predictions [5-11], and drug-drug interaction (DDI) predictions [12-13]. A key advantage behind these methods is that deep learning algorithms can capture the complex nonlinear relationships between input and output data [14].

In the past few years, deep learning techniques have gradually emerged as a powerful paradigm for drug discovery. Most deep learning architectures, such as convolutional neural networks [15] and recurrent neural networks [16], operate only on regular grid-like data (e.g., 2D images and text sequences), and are not well suited for graph data (e.g., DDI and DTI networks). However, in the real world, biomedical data are often formed as graphs or networks. In particular, biomedical heterogeneous networks (BioHNs) that integrate multiple types of data source are used extensively for life science researches. This is intuitive since BioHNs are well suited for modeling complex interactions in biological systems. For example, the BioHNs incorporating DDIs, DTIs, and protein-protein interactions (PPIs), protein-disease associations can naturally simulate the 'multi-drug, multi-target, multi-disease' biological processes within human body [17]. In the context of biomedical networks applications, graph neural networks (GNNs) [18-20], which are deep learning architectures specifically designed for graph structure data, are utilized to improve drug discovery. Such studies [21-24] use GNNs to generate the representation of each node in BioHNs, and formulate drug discovery as the node- or edge-level prediction problems. These graph neural network-based drug discovery approaches have shown high-precision predictions. However, most existing methods heavily depend on the size of training samples; that is, only large-scale training samples can help models to achieve great performance. Concurrently, with the variation of training sample sizes, the performance is changed by a large margin. Unfortunately, data labeling is expensive and time-consuming. Therefore, these graph-based deep learning models that rely on large-scale labeled data may not be satisfactory in real drug development scenarios.

Self-supervised representation learning (SSL) is a promising paradigm for solving the above issues. In SSL, deep learning models are trained via pretext tasks, in which supervision signals are automatically extracted from unlabeled data without the need for manual annotation. SSL aims to guide models to generate the generalized representations to achieve better performance on various downstream tasks. Following the immense success of SSL on computer vision [25-26] and natural language processing [27-28], SSL models built upon BioHNs have enjoyed increasing attention and have been successfully applied to drug discovery [29-32]. Unfortunately, most existing methods often design a single SSL task to train GNNs for drug discovery, thus leading to the built-in bias toward a single task and ignoring the multi-perspective characteristics of BioHNs. To cope with the potential bottleneck in single task-driven SSL applications, there are a few attempts leveraging multiple SSL tasks for facilitating performance of drug discovery [33-35]. These methods aim to integrate the advantages of various types of SSL tasks via the multi-task learning paradigms. However, most previous approaches train GNNs according to a fixed joint strategy involving multiple tasks, and do not focus on the differences between various multi-task combinations. Concurrently, the determination of which combination strategies can generate the most effective improvements has rarely been explored. Therefore, it is significant to pay attention to the choice of multi-task combination strategies in SSL approaches. In addition, multi-task SSL methods built on BioHNs for drug discovery are still in the initial stages, and more systematic studies are pressingly needed.

To address the aforementioned problems, we propose multi-task joint strategies of self-supervised representation learning on biomedical networks for drug discovery, named MSSL2drug. Inspired by three modality features (i.e., structures, semantics, and attributes in BioHNs), six self-supervised tasks are developed to explore the impact of various SSL models on drug discovery. Next, fifteen multi-task joint strategies are evaluated via a graph attention-based multi-task adversarial learning model in two drug discovery scenarios. We find that the combinations of multimodal tasks can generate best performance compared to other multi-task strategies. Another interesting conclusion is that the local-global combination models tends to yield good results than random task combinations when there are the same sizes of modalities.

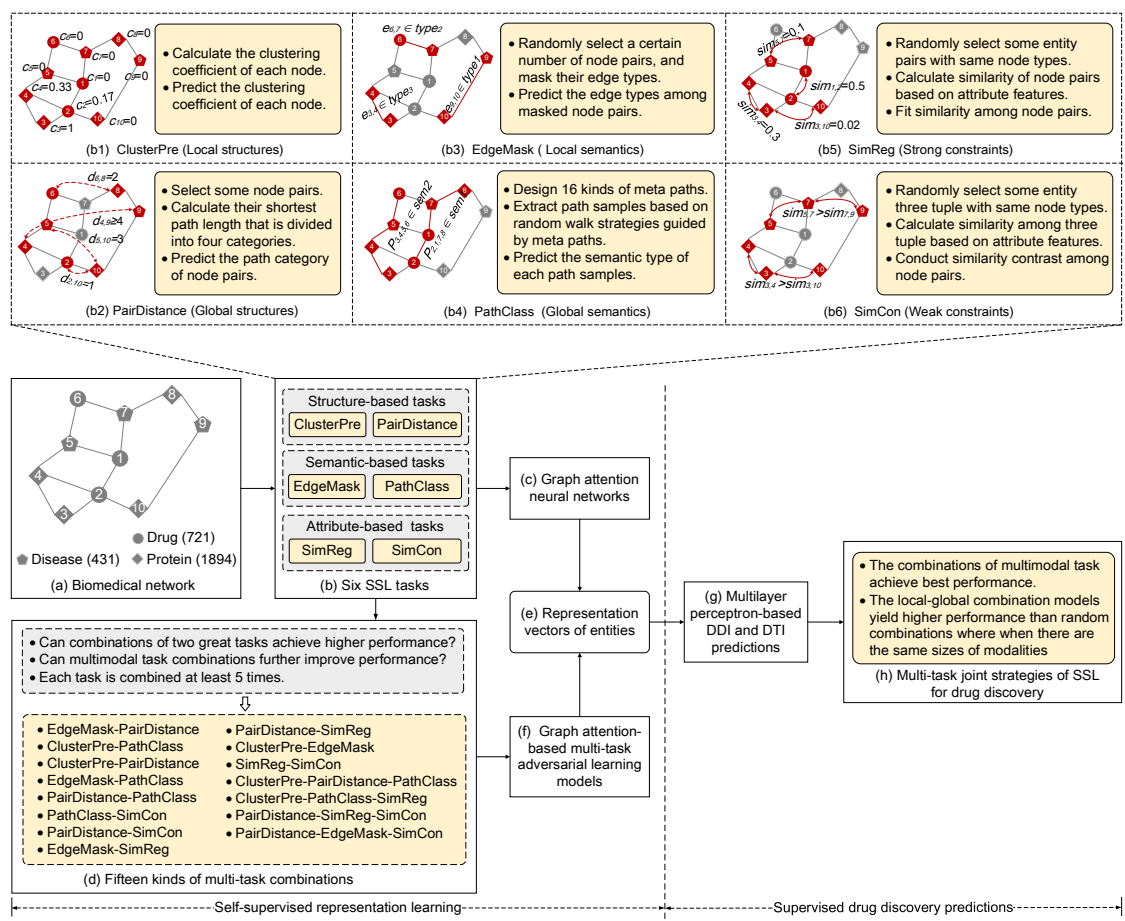

**Figure 1.** The schematic workflow of MSSL2drug. All circles, quadrangles, and pentagons denote the drugs, proteins, and diseases in a BioHN, respectively. The solid lines are the relationships among the biomedical entities in a BioHN. The red nodes represent the randomly selected vertices or node pairs in each of self-supervised task. The red solid lines in the edge type masked prediction

(EdgeMask) and bio-path classification (PathClass) modules represent the randomly selected edges or paths. The dashed curved lines in the pairwise distance classification (PairDistance) module represent the measurements of the shortest paths between biomedical entities. The dashed solid lines in the node similarity regression (SimReg) and node similarity contrast (SimCon) modules represent the measurements of the similarities between biomedical entities. ClusterPre and PairDistance denotes clustering coefficient prediction and a pairwise distance classification, respectively.

## 2 Result

### 2.1 Overview of MSSL2drug

As shown in Fig. 1, we propose the multi-task joint strategies of self-supervised representation learning on biomedical networks for drug discovery, named MSSL2drug. First, we construct a biomedical heterogeneous network that integrates 3,046 biomedical entities and 111,776 relationships. Second, we develop six self-supervised tasks based on structures, semantics, and attributes in the BioHN, as shown in Fig. 1(b). These self-supervised tasks guide graph attention networks (GATs) to generate the representations from different views in the BioHN. More importantly, we develop fifteen kinds of multi-task combinations and a graph attention-based multi-task adversarial learning framework to improve the representation quality. Finally, the different representations from single-task and multi-task SSL are fed into multilayer perceptron (MLP) for predicting DDIs and DTIs. Based on the experiment results, we can draw two important findings. (1) The combinations of multimodal SSL tasks achieve state-of-the-art performance of drug discovery. (2) The joint training of local and global SSL tasks is superior to the random combinations of two SSL tasks when there are the same sizes of modalities.

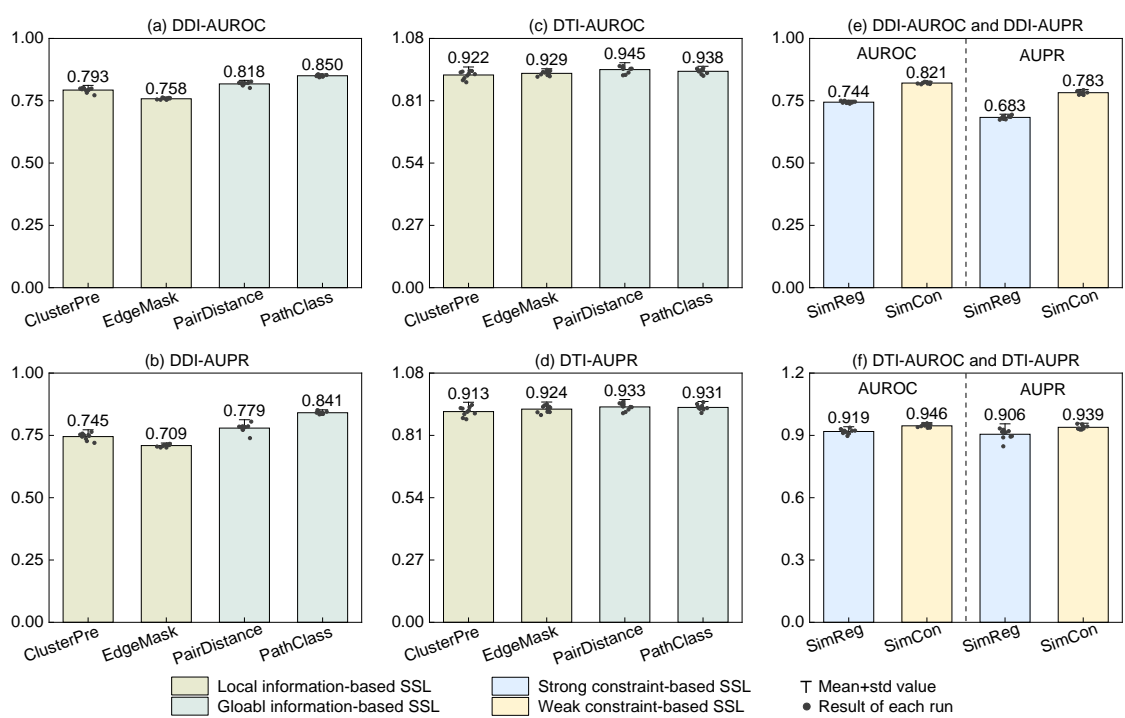

**Figure 2.** The results of single SSL tasks for drug warm start predictions, where the area under precision recall (AUPR) curve and area under receiver operating characteristic (AUROC) curve are used for the evaluation metrics. The mean and std values denote average and standard deviation values that are calculated across ten results.

## 2.2 Performance of single task-driven SSL

In single task-driven SSL on drug warm start predictions, PairDistance and PathClass achieve relatively higher results, as shown in Fig. 2. Based on Student's t-test on the DTI and DDI results (as described in Supplementary Material Section S1), we find that PairDistance and PathClass significantly outperform ClusterPre and EdgeMask ($p$-value $<0.05$). Another aspect to note is that SimCon is superior to SimReg. These results suggest that the global information-driven SSL approaches are superior to the local information-based SSL. Previous study [36] also made a similar finding. In addition, attribute weak constraint-based SSL tasks outperform strong constraint-based models.

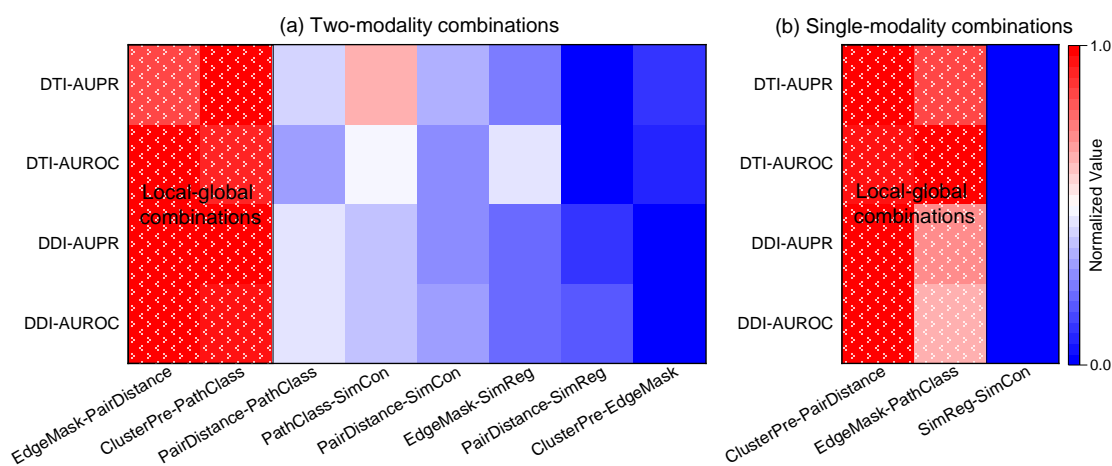

**Figure 3.** Heatmap of two-task combinations for drug warm start predictions where the results are normalized to [0,1] along the x-axis by the Min-Max normalization technique. The redder (bluer) squares denote the greater (smaller) the value. The shaded area denotes the combinations of global and local SSL tasks.

## 2.3 Local-global tasks achieve superior performance

In this experiments, first, 11 two-task combination models are divided into two categories: single-modality combinations and two-modality combinations. It is noted that we design self-supervised tasks inspired by various modality knowledge including structures, semantics, and attributes in BioHNs. Therefore, there are up to three single-modality combination models, as shown in Fig. 3(b). Second, we compare the performance of two-task models with same size of modalities. The results in Fig. 3 suggest that the joint training of local and global SSL tasks (i.e., EdgeMask-PairDistance, ClusterPre-PathClass, ClusterPre-PairDistance and EdgeMask-PathClass) tends to obtain higher performance than random combinations of two SSL tasks when there are the same sizes of modalities. To further investigate the difference among various methods, we provide more analyses and a Student's t-test in Supplementary

Material Section S2. Therefore, we conjecture that the local-global combination strategies can be regarded as an effective guideline for multi-task SSL to drug discovery.

**2.4 Multimodal tasks achieve best performance**

The results in Fig. 4 show an interesting situation; that is, the growth of modalities leads to the significant performance improvement ($p$-value $<0.05$) for drug discovery. The more results and Student's t-test analyses can be found in Supplementary Material Section S3. These results suggest that combinations of multimodal tasks can achieve best performance for drug discovery. Therefore, we conjecture that the multimodal combination strategy can be regarded as a potential guideline for multi-task SSL for drug discovery.

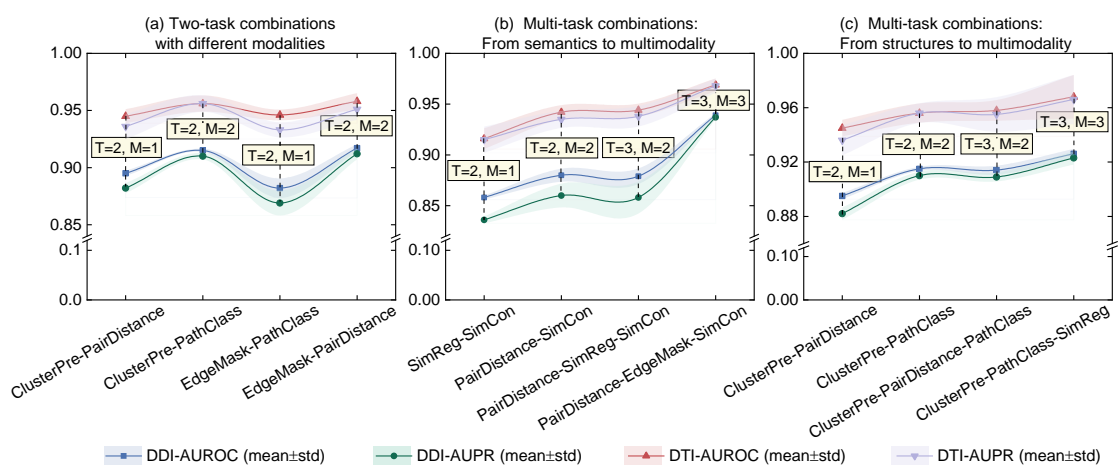

**Figure 4.** The results obtained multimodal task combinations for drug predictions, where 'T' and 'M' denote the total number of tasks and modalities in each combination, respectively. The mean and std values denote the average and standard deviation values that are calculated across ten results.

**2.5 Performance of MSSL2drug on cold start predictions**

For the cold start drug prediction scenarios, the results of DDI and DTI predictions are generated by six basic SSL tasks and fifteen kinds of multi-task combinations. These results are straightforward and effective demonstrations that global information and attribute weak constraint-based SSL models can achieve better performance than local information and attribute strong constraint-based SSL. More importantly, these results verify that multimodal and local-global combination strategies can achieve state-of-the-art the prediction performance of drug discovery. The detailed result analyses can be found in Supplementary Material Section S4.

## 3 Discussion

Recently, self-supervised representation learning on biomedical heterogeneous networks has emerged as a promising paradigm for drug discovery. Therefore, we aim to explore a combination strategy of multi-task self-supervised learning on biomedical heterogeneous networks for drug discovery. Based on six self-supervised learning tasks, we find that global knowledge-based SSL models outperform local information-based SSL models for drug discovery. This is intuitive and understandable since global view-based SSL tasks can capture the complex structures and semantics that is unable to be naturally learned by local SSL models. We also find that attribute weak constraint-based SSL tasks are superior to strong constraint-based models. This may be attributed to the fact that the similarity scoring functions are handcrafted and unable to accurately reflect the similarities among nodes in the original feature space. Unfortunately, the node similarity regression tasks arbitrarily fit node similarity values of node pairs. In contrast, similarity contrast tasks reduce the dependence on the original feature similarity values.

More importantly, fifteen kinds of multiple task combinations are evaluated by a graph attention-based multi-task adversarial learning model for drug discovery. These results suggest that the joint training the global and local tasks can achieve the relatively high prediction performance when there are the same sizes of modalities. In contrast, combining the tasks with great performance does not necessarily lead to better performance than other multi-task combinations for drug discovery. This is intuitive since there may be some conflicts and redundancies in the random combinations of SSL tasks. However, the combinations of global and local SSL models enable GNNs to leverage complementary information in BioHNs. To be specific, the local graph SSL models can capture the features within node itself or its first-order neighbors, but ignore the bird's-eye view of the node position in BioHNs. Fortunately, global SSL models can learn the dependencies among long-range neighborhoods, thus compensating the shortcomings of local SSL tasks.

Simultaneously, an interesting finding is that combination models with multimodal tasks tend to generate best performance. This is because the combinations of multimodal tasks can capture multi-view information including structure, semantic and attribute features in BioHNs. The multimodal SSL models allow for knowledge transfer across multiple views and attain a deep understanding of natural phenomena in BioHNs. For a given SSL task, there are different levels of contributions in different multi-task combinations. Generally, if a SSL task can bring new modality information to multi-task models, it will generate the relatively greater contributions. In addition, if a local (global) information-driven SSL task is added to global (local) information-driven SSL tasks, it tends to bring high performance improvement. The multimodal and local-global combination strategies may be prioritized, when developing multi-task SSL for drug discovery. In other words, you can design yourself multi-task SSL models according to the multimodal and local-global combination strategies when you want to use MSSL2drug for drug discovery. On the other hand, you can also directly use PairDistance-EdgeMask-SimCon for drug discovery, because it integrates the multimodal and local-global SSL tasks, and achieves best performance.

In conclusion, self-supervised representation learning based on biomedical heterogeneous networks provides new opportunities for drug discovery. To facilitate this line of researches, we carefully explore the influence of various basic SSL tasks and propose unified combination strategies involving multi-task self-supervised representation learning to improve drug discovery. Simultaneously, we present a detailed empirical study to understand which combination strategies of multiple SSL tasks are most effective for drug discovery. In the future, we will pay attention to designing more SSL tasks and combination strategies to further improve performance of drug discovery. In addition, multi-class DDI predictions are closer to real-world drug discovery. Nevertheless, it is more challenging and difficult to manually annotate multi-class DDI data. Therefore, we will further verify the proposed global-local and multimodal combination strategies on multi-class DDI predictions.

## 4 Materials and methods

### 4.1 Biomedical heterogeneous networks

In this work, we construct a biomedical heterogeneous network (BioHN) according to deepDTnet [37]. The constructed BioHN assembles three types of nodes (i.e., drugs, proteins and diseases) and five types of edges (drug-drug interactions, drug-protein interactions, drug-disease associations, protein-protein interactions, and protein-disease associations). More specifically, the drug-drug interactions are extracted from the DrugBank database (v4.3) [38], where we only select drugs that have experimentally validated target information. The chemical name of each drug is transferred to a DrugBank ID. The drug-protein interaction networks are collected from the DrugBank database (v4.3), PharmGKB [39], and the Therapeutic Target database (TTD) [40]. We extract human protein-protein interactions with multiple pieces of evidences from the HPRD database (Release 9) [41], HuRI [42] and BioGRID [43]. Each protein name is transferred into an Entrez ID (https://www.ncbi.nlm.nih.gov/gene) via the NCBI (https://www.ncbi.nlm.nih.gov/). Drug-disease associations are attained via the fusion of the drug indications in the repoDB [44], DrugBank (v4.3), and DrugCentral databases [45]. Disease-protein associations are collected from two databases, including the Online Mendelian Inheritance in Man (OMIM) database [46] and the Comparative Toxicogenomics database (CTD) [47]. The disease names are standardized according to Unified Medical Language System (UMLS) vocabularies [48], and mapped to the MedGen ID (https://www.ncbi.nlm.nih.gov/medgen/) based NCBI database. BioHN in this work includes less information profiles than the dataset deepDTnet. Finally, the BioHN contains 3,046 nodes and 111,776 relationships (as described in Supplementary Material Table S7). There are 1,894 proteins, 721 drugs, and 4,978 drug-protein interactions in the BioHN. The ratio of DTI labels is 0.003≈4,978/(721*1,894). Similarly, there are 66,384 drug-drug interactions in the BioHN. Therefore, the ratio of DDI label is 0.256≈66,384/(721*720*0.5). In other words, there are sparse labels for DDI and DTI predictions. Therefore, we propose MSSL2drug that explore multi-task joint strategies of self-supervised representation learning on biomedical networks for drug discovery.

## 4.2 Basic self-supervised learning tasks

Multimodal information, including structures, semantics, and attributes in BioHNs, provides unprecedented opportunities for designing advanced self-supervised pretext tasks. Hence, we develop six self-supervised tasks based upon the multimodal information contained in BioHNs for drug discovery.

### 4.2.1 Structure-based SSL tasks

The first direct choice for constructing SSL tasks is the inherent structure information contained in BioHNs. For a given node, self-supervision information is not only limited to itself or local neighbors, but also includes a bird's-eye view of the node positions in a BioHN. Therefore, we design a clustering coefficient prediction (ClusterPre) task that captures local structures and a pairwise distance classification (PairDistance) task that reflects the global structure information in BioHNs.

**Clustering coefficient prediction (ClusterPre):** In this pretext task, we use GATs to predict the clustering coefficient [49] of each node in BioHNs. The ClusterPre SSL task aims to guide GATs to generate low-dimensional representations that preserve the local structure information in BioHNs. In ClusterPre, the loss function adopts the mean squared error (as described in Supplementary Material Section S5.1).

**Pairwise distance classification (PairDistance):** We develop PairDistance that is not limited to a node itself and its local neighborhoods; it also takes global views of a BioHN. Similar to $S^2$GRL [50], we randomly select a certain number of node pairs and calculate the shortest path length between each node pair $(i, j)$ as its distance value $d_{i,j}$. Subsequently, these node pairs and distance values are used to train GATs for drug discovery. In practice, the distances between node pairs are divided into four categories, that is, $d_{i,j}=1, d_{i,j}=2, d_{i,j}=3$ and $d_{i,j} \geq 4$. In other words, the PairDistance SSL task can be treated as a multiclass classification problem in which we adopt the cross entropy loss function (as described in the Supplementary Material Section S5.2). This is mainly attributed to two reasons. (1) The distinctions between the node pairs interacting via longer paths (i.e., $d_{i,j} \geq 4$) are relatively vague; thus, it is more reasonable to divide the longer pairwise distances into one "major" class [50]. (2) Based on the small-world phenomenon [51], we suppose that the shortest path lengths between most node pairs are within a certain range (as described in Supplementary Material Section S5.2). If we fit longer pairwise distances, some noisy values will be generated. Here, $d_{i,j} \geq 4$ indicates that PairDistance is not limited to the local connections in BioHNs. Therefore, PairDistance is beneficial for guiding GATs to generate node representations that encode the global topology of BioHNs. In addition, node pairs via random selection may lead to unstable results in PairDistance. Therefore, we repeat this process numerous times, and then the average performance is computed.

#### 4.2.2 Semantic-based SSL tasks

BioHNs integrate multiple types of nodes or edges. The different relationships among these nodes contain distinct semantic information. Recent studies have suggested that semantic information can contribute to learning high-quality representations [28, 31]. Therefore, we develop edge type masked prediction (EdgeMask) task and bio-path classification (PathClass) task for encouraging GATs to capture certain aspects of semantic knowledge. Similar to the structure-based SSL tasks, EdgeMask and PathClass can capture the local and global semantics of BioHNs, respectively.

**Edge type masked predictions (EdgeMask):** This task is inspired by the BERT model [27], in which the core is a masked language model [52]. More specifically, we randomly mask edge types among some node pairs and then use GATs to predict these edge types, where the edge representation vectors are obtained by concatenating the representations of their two end-nodes. A detailed description of EdgeMask is found in Supplementary Material Section 5.3. The types of edges indicate the different action mechanisms between biomedical entities. Therefore, EdgeMask can enable GATs to learn the semantic features among local neighborhoods.

**Bio-meta path classifications (PathClass):** Compared to the types of edges among nodes, meta paths are a sequences for incorporating the complex semantic relationships in BioHNs (as described in the Supplementary Material Section S5.4). Different types of meta paths indicate distinct semantics. In PathClass, we design 16 types of meta paths as shown in the Supplementary Material Table S8, where the first or last objects are drugs or proteins, respectively. This is mainly because drugs and proteins are interconnected with other entities by more edges (as described in Supplementary Material Table S9). These meta paths guide random walks to extract path samples from BioHNs. In addition, we generate an equal number of false path instances by randomly replacing some nodes in true path instances. To be specific, for a given true path instance, it has 6.25% (i.e., 1/16) chance being replaced to generate a false path instance (as described in Supplementary Material Section S5.4). Therefore, all path samples are divided into 17 categories, including 16 kinds of true meta paths and one kind of false meta paths. Finally, we use GATs to predict the type of each path sample for learning node representations that contain rich semantics and complex relationships. Similarly, we adopt the cross entropy as loss function in PathClass.

#### 4.2.3 Attribute-based SSL tasks

In addition to structures and semantics, attribute features play key roles in self-supervised representation learning. More generally, nodes with similar properties, such as the simplified molecular input line entry system (SMILES) strings [53] of drugs, should be distributed closely in the representation space. However, GATs only aggregate the features of node itself and its local neighborhoods, thus losing the similarity features among nodes. Based on this intuition, we develop two attribute-based SSL tasks, i.e., node similarity regression (SimReg) and node similarity contrast (SimCon), to enable GATs to maintain the similarity attributes in the original feature space. According to the degree of dependence on the original feature similarities, SimReg and SimCon can be categorized as strong constraint- and weak constraint-based SSL paradigms, respectively.

**Node similarity regression (SimReg):** The proposed SimReg task requires GATs to fit similarity values of node pairs. More specifically, we randomly select a certain number of node pairs $(i, j)$ (where $i$ and $j$ are the same types of nodes); and then calculate their similarity value $sim_{i,j}$ in the original feature space, such as the similarity between drug SMILES sequences. We require GATs to fit the similarity values ($sim_{i,j}$) of node pairs in the original feature space as possible. In other words, SimReg encourages GATs to learn representations via a strong constraint-based SSL paradigm. In this work, we use different property similarity measurements in the various types of nodes. The Tanimoto coefficient [54] among the SMILES sequences of drugs are treated as drug-drug similarity scores. We leverage the Smith-Waterman algorithm [55] to calculate the sequence similarity scores of protein pairs. The disease similarity scores are obtained by using PPI-based ModuleSim algorithm [56]. The detailed similarity measurement approaches and objective functions are described in Supplementary Material Section 5.5.

**Node similarity contrast (SimCon):** In SimReg, the similarity scoring mechanisms have an important impact on the representation learning process. SimReg cannot guarantee to generate the high-quality representations when similarity scores may not accurately reflect the true similarity values among nodes in original feature space. Therefore, we propose SimCon to reduce the influence of similarity scoring mechanisms. In SimCon, it assumes that the similar nodes in the original feature space should be closer in the embedding space than dissimilar nodes. More specifically, we randomly select a certain number of three tuples $(i, j, k)$ for nodes, where $i$, $j$ and $k$ belong to the same types of nodes and $sim_{i,j} \geq sim_{i,k}$. For a given tuple $(i, j, k)$, we use GATs to conduct a node similarity contrast; that is, the cosine values ($\cos_{i,j}$ and $\cos_{i,k}$) between the node representations generated by GATs should satisfy $\cos_{i,j} \geq \cos_{i,k}$. Formally, we propose a novel objective function:

$$\ell_{simCon}(\theta) = \frac{1}{|M|} \sum_{(i,j,k) \in M} L(i, j, k) \qquad (1)$$

where $M$ is the selected set of three tuples $(i, j, k)$, $|M|$ is the number of three tuple, and $L(i, j, k)$ is calculated as follows:

$$L(i, j, k) = \begin{cases} 0, & \cos(f_\theta(i), f_\theta(j)) - \cos(f_\theta(i), f_\theta(k)) \geq 0 \\ g(i, j, k), & \text{otherwise} \end{cases} \quad (2)$$

where $g(i, j, k)$ is calculated as follows:

$$g(i, j, k) = sim_{i,j} - sim_{i,k} - \left(\cos(f_\theta(i), f_\theta(j)) - \cos(f_\theta(i), f_\theta(k))\right) \quad (3)$$

where $\theta$ is the parameters of a graph neural network $f_\theta(\cdot)$ and $f_\theta(i)$ denotes the embedding vectors of node $i$. In addition, $\cos(\cdot,\cdot)$ is the cosine similarity value between two embedding vectors.

Obviously, SimCon only requires that GATs can distinguish the similarity distributions between node pairs $(i, j)$ and node pairs $(i, k)$. However, SimReg requires that GATs fit similarity values for node pairs. Therefore, SimCon reduces the dependence on the original feature similarity values compared to SimReg; thus, SimCon is a weak constraint-based SSL paradigm.

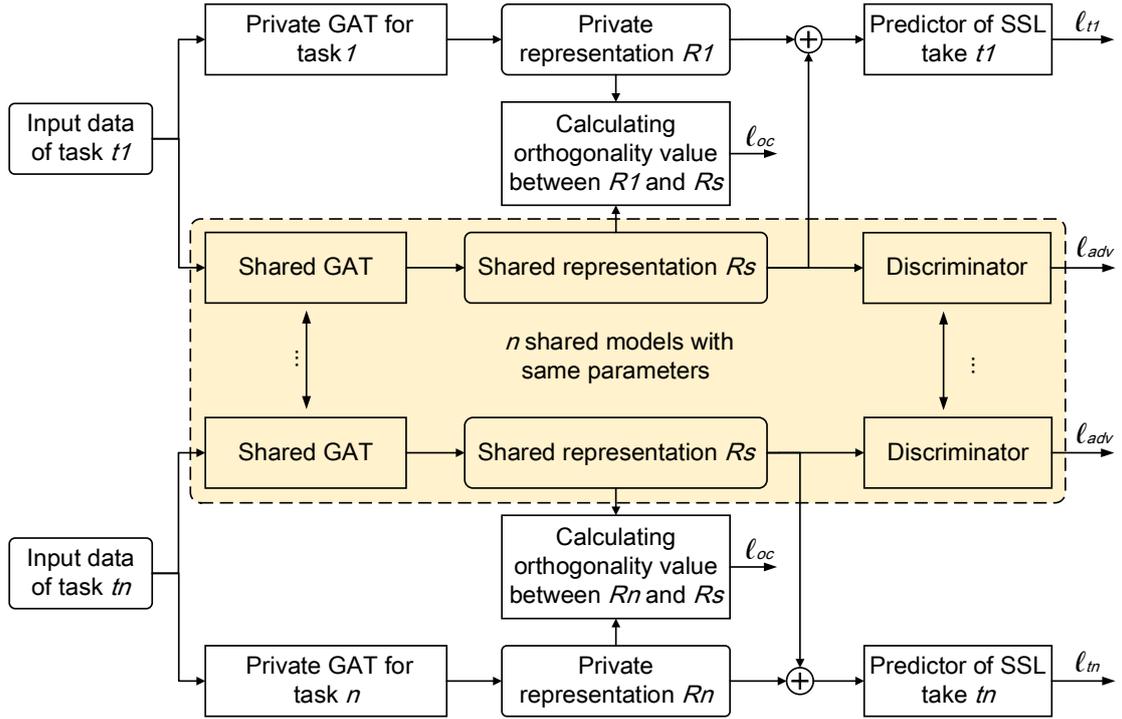

**Figure 5.** The framework of graph attention-based adversarial multi-task learning. For each epoch, we randomly select a SSL task *tn* from multi-task combinations. The corresponding private and shared GAT models generate the task-specific representations (*Rn*) and common representations (*Rs*), respectively. The *Rn* and *Rs* are concatenated, and then fed into the MLP-based predictor of SSL task *tn*. In addition, the *Rs* are fed into the MLP-based discriminator to predict what kind of task the shared representation vectors come from. The parameters of current private and shared GAT models are updated by back-propagation based on the loss values from a SSL task predictor and discriminator, respectively. Finally, the parameters of the current shared model are assigned to all the other shared models. Therefore, we attain *n* private GAT models and shared GAT models with same parameters after multi-task SSL training. In other words, MSSL2drug generates the private representations by all private GATs and the shared representation by an arbitrary shared model.

### 4.3 Graph attention-based multi-task adversarial learning

In this work, the integration of the multi-task learning and GATs is a challenging and critical problem. Inspired by [57], we propose a graph attention-based adversarial multi-task learning framework for drug discovery, as shown in Fig. 5. The graph attention-based multi-task adversarial learning framework can be divided into the private and share parts that employ graph attention networks (GATs) [19] with different parameters.

### 4.3.1 Graph attention network

The graph attention network (GAT) is a popular graph neural network. GAT assumes that the contributions of neighboring nodes to the central nodes are different. To calculate the representations of one node, GAT aggregates its neighbor features by a multi-head attention mechanism. For a given node, the features from multiple attention mechanism models are concatenated to generate the final representation vectors. The final output features of each node can be calculated by:

$$\vec{h}_i' = \underset{k=1}{\overset{K}{\|}} \sigma\left(\sum_{j \in N_i} \alpha_{ij}^k \mathbf{W}^k \vec{h}_j\right) \quad (4)$$

where $\sigma$ is a nonlinearity activation function, $K$ is the number independent attention mechanisms, $\mathbf{W}^k$ is the weight matrix of linear transformation in the *k*-th attention mechanism, $N_i$ is the number of neighbors of node $i$, $\|$ represents concatenation operation, $\vec{h}_j$ is the current representations of neighbor $j$. More importantly, $\alpha_{ij}^k$ is the attention coefficients computed by the *k*-th attention mechanism. Intuitively, there are $K$ attention coefficients between node $i$ and $j$. $\alpha_{ij}^k$ can be calculated by:

$$\alpha_{ij}^k = \frac{\exp\left(\text{LeakyReLU}\left((\vec{\mathbf{a}}^k)^\mathrm{T}\left[\mathbf{W}^k \vec{h}_i \| \mathbf{W}^k \vec{h}_j\right]\right)\right)}{\sum_{j \in N_i} \exp\left(\text{LeakyReLU}\left((\vec{\mathbf{a}}^k)^\mathrm{T}\left[\mathbf{W}^k \vec{h}_i \| \mathbf{W}^k \vec{h}_j\right]\right)\right)} \quad (5)$$

where $\vec{\mathbf{a}}^k$ is a weight vector in *k*-th attention mechanism, $(\cdot)^\mathrm{T}$ represents transposition.

### 4.3.2 Task discriminator

For any node $i$ in task $t$, the shared GAT generates task-invariant representations $x_t^i = f_{\theta_s}(i)$ where $\theta_s$ is the parameter of the shared GAT $f_{\theta_s}(\cdot)$. Then, these representation vectors $x_t^i$ are fed into a multilayer perceptron that is treated as a task discriminator. This multilayer perceptron aims to predict what kind of task the shared representation vectors come from.

$$D(x_t^i, \theta_{td}) = \text{softmax}\left(\text{MLP}_{\theta_{td}}(x_t^i)\right) \tag{6}$$

where $\text{MLP}(\cdot)$ is a multilayer perceptron in which the trainable parameter is $\theta_{td}$.

The loss values from the task discriminator can be calculated as follows:

$$\ell_{adv} = \min_{\theta_s}\left(\max_{\theta_{td}} \sum_{t=1}^{T}\sum_{i=1}^{N_t} y_t^i \log\left[D\left(f_{\theta_s}(i), \theta_{td}\right)\right]\right) \tag{7}$$

where $N_t$ is the number of training nodes in task $t$, and $y_t^i$ denotes the ground-truth labels indicating the type of current task.

### 4.3.3 Orthogonality constraints

The above shared model generates some features that may appear in both the shared space and the private space. Therefore, we adopt an orthogonality constraint [57] to eliminate redundant features from the private and shared spaces. Formally, the objective function of the orthogonality constraint is calculated as follows:

$$\ell_{oc} = \sum_{t=1}^{T}\sum_{i=1}^{N_t} \left\| f_{\theta_t}(i)^{\text{T}} \cdot f_{\theta_s}(i) \right\|_F^2 \tag{8}$$

where $\|\cdot\|_F^2$ is the squared Frobenius norm, and $f_{\theta_t}(\cdot)$ is the private GAT of the current task $t$.

### 4.3.4 Multi-task adversarial training

The final loss function of multi-task SSL can be written as follows:

$$\ell_{total} = \ell_t + \lambda \ell_{adv} + \gamma \ell_{oc} \tag{9}$$

where $\lambda$ and $\gamma$ are hyperparameters. $\ell_t$ denotes the loss value of task $t$.

During multi-task learning phase, inspired by [58], the models are trained in a stochastic manner by looping over the tasks.

***Step 1*:** Randomly select a task.

***Step 2*:** Sample an epoch of instances from the task and train the corresponding private model and shared model.

***Step 3*:** Update the corresponding parameters by back-propagation. Subsequently, the parameters of the current shared model are assigned to all the other shared models.

***Step 4*:** Go to Step 1.

In this way, multiple private and shared GAT models are updated by the corresponding specific task. However, in practice, these shared GAT models are equivalent to a GAT model because they have the same parameters. In other words, we attain multiple private GAT models and a shared GAT model. Therefore, in two-task learning cases, Fig. S4(a) is equivalent to Fig. S4(b) in Supplementary Materials.

For a given node, different SSL tasks in different epochs guide the shared GAT to capture the features with itself task property. Therefore, self-supervised training in different epochs can be treated as the adversarial learning process, that is, each SSL task encourages shared GAT to generate task-specific representations. After sufficient training, the shared GATs reach a point, at which it integrates the property of different tasks. Therefore, the shared feature space simply contains common information. In contrast, the private GAT model generates task-specific representations to make accurate SSL predictions.

**4.4 Initialization features**

In MSSL2drug, the initialization features of each node and adjacency matrixes of BioHNs are fed into GATs to perform training and test. Here, we take an example to describe the process of feature initialization, as shown in Supplementary Material Fig. S5. There are three key steps to generate the initialization features. For each given node, its neighbors are divided into three categories (i.e., drugs, proteins, and diseases).

**Step 1:** Counting the number of neighbors in each class, $X = \{x_1, x_2, \cdots, x_N\}$, $Y = \{y_1, y_2, \cdots, y_N\}$ and $Z = \{z_1, z_2, \cdots, z_N\}$, respectively, where $N$ is the total number of nodes. For instance, for given *node 1*, $x_1=1$, $y_1=2$, $z_1=1$, the sum of $x_1, y_1, z_1$ is its degree (i.e., the number of its neighbors), as shown 1$^{st}$ row in Supplementary Material Fig. S5(b);

**Step 2:** Converting $X$, $Y$ and $Z$ to matrixes $\mathbf{X} = \{\vec{u}_1, \vec{u}_2, \cdots, \vec{u}_N\}$, $\mathbf{Y} = \{\vec{v}_1, \vec{v}_2, \cdots, \vec{v}_N\}$, and $\mathbf{Z} = \{\vec{g}_1, \vec{g}_2, \cdots, \vec{g}_N\}$ by one-hot encoding technologies (https://www.educative.io/blog/one-hot-encoding);

**Step 3:** Generating initialization feature matrix $\mathbf{F} = \{\vec{u}_1 \| \vec{v}_1 \| \vec{g}_1, \vec{u}_2 \| \vec{v}_2 \| \vec{g}_2, \cdots, \vec{u}_N \| \vec{v}_N \| \vec{g}_N\}$ by concatenating $\mathbf{X}$, $\mathbf{Y}$, and $\mathbf{Z}$, where $\|$ is a concatenation operation.

**4.5. Experiment settings**

**4.5.1 Multi-task combination settings**

We design various multi-task combinations to answer two key questions.

- **Can joint training of two tasks with great performance (like 'Alliance between Giants') achieve higher performance than random combination of two tasks?**

  The results of single task SSL suggest that PairDistance, PathClass, and SimCon achieve the relatively higher performance. Therefore, we first chose all combinations of 'Alliance between Giants' (i.e., PairDistance-PathClass, PathClass-SimCon, and PairDistance-SimCon). Next, we randomly select 8 other two-task combinations, i.e., EdgeMask-PairDistance, ClusterPre-PathClass, ClusterPre-PairDistance, EdgeMask-PathClass, EdgeMask-SimReg, PairDistance-SimReg, ClusterPre-EdgeMask, SimReg-SimCon.

- **Can the combinations integrating multimodal information further improve the prediction performance?**

  Based on 11 two-task combinations, we select four multi-task combinations to evaluate the influence of different modalities. As shown in Table S10 in Supplementary Materials, there is only one different task in context compositions. For example, PairDistance-SimCon is turned into PairDistance-EdgeMask-SimCon by adding SimReg. In addition, the pool of combination strategies keep diversity criterions, that is, each task is combined at least 5 times. Therefore, we selects 15 kinds of task combinations to guaranteed reliability.

### 4.5.2 Drug discovery predictions under different scenarios

In this study, we focus on the performance of various SSL tasks on DDI and DTI predictions, because they are key stages and play important roles in various applications of drug discovery. Simultaneously, DDI and DTI predictions are treated as link predictions in homogeneous and heterogeneous networks, respectively. Therefore, DDI and DTI predictions can systematically demonstrate the performance of various kinds of SSL tasks and combination strategies. According to the guidance of KGE_NFM [5], we design the following two experimental scenarios. **Warm start predictions:** Given a set of drugs and their known DTIs, we aim to predict other potential interactions between these drugs. All the known interactions are positive samples, and an equal number of negative samples are randomly selected from the unknown interactions. The positive and negative samples are split into a training set (90%) and a testing set (10%). In this situation, the training set may include drugs and targets contained in the test set. The same experimental setting as DTI predictions are used for DDI predictions. In this experimental scenarios, we compare the differences among various SSL tasks for DDI and DTI predictions, and draw a conclusion on which combination strategies can generate the best performance. **Cold start for drugs:** In real drug discovery, it is more important and challenging to predict potential targets and drugs that may interact with newly discovered chemical compounds. In other words, the test set contains drugs that are unseen in the training set. To be specific, we randomly select 5% drugs, and then all DTI and DDI pairs associated with these drugs are treated as test set. This scenario aims to validate the conclusions that are found in the warm start predictions. We use the area under precision recall (AUPR) curve and area under receiver operating characteristic (AUROC) curve as the evaluation metrics for drug discovery. To reduce the data bias and uncertain disturbance, each model is executed 10 times, and the average performance is computed. The hyperparameter selections can be found in Supplementary Material Section S6.

## Data availability

All relevant data including the original network and initialization features can be downloaded from https://github.com/pengsl-lab/MSSL.git.

## Code availability

The source code can be found at: https://github.com/pengsl-lab/MSSL.git. In the GitHub repository, we have provided source code that include the data processing of six SSL pretext tasks, GAT-based multi-task representation models, and MLP-based DDI or DTI predictors. Concurrently, we added the description of how to use program.

## Acknowledgement

This work was supported by NSFC Grants: U19A2067 and 81973244; National Key R&D Program of China 2022YFC3400404; Science Foundation for Distinguished Young Scholars of Hunan Province: 2020JJ2009; Science Foundation of Changsha: Z202069420652, kq2004010; JZ20195242029, JH20199142034; The Funds of Strategic Priority Research Program of Chinese Academy of Sciences: XDB38040100; The Funds of State Key Laboratory of Chemo/Biosensing, and Chemometrics, and the Cloud Brain and Major Key Project of Peng Cheng Lab.

## Author contributions

X.W. and Y.C. conceived the original idea and developed the code for the core algorithm. S. P designed the experiment and wrote the initial version of the manuscript. F.L. and Y.Y analyzed the experimental data and edited this manuscript. Y.N.Y constructed the biomedical network data. All authors reviewed and approved the final manuscript.

**Competing interests:** The authors declare no competing interests.

Supplementary Materials for

# Multi-task Joint Strategies of Self-Supervised Representation Learning on Biomedical Networks for Drug Discovery

Xiaoqi Wang[†], Yingjie Cheng[†], Yaning Yang, Yue Yu,
Fei Li[*], and Shaoliang Peng[*]

[†]These authors contributed equally to this work.

*Corresponding author. Email:

pittacus@gmail.com (Fei Li);

and slpeng@hnu.edu.cn (Shaoliang Peng).

# The Supplementary Materials file includes：

## S1. Result analysis of single task-driven SSL on drug warm start predictions

PathClass attains approximately 10-15% improvements over EdgeMask in terms of AUROC and AUPR scores for DDI and DTI predictions. Another aspect to note is that SimCon is superior to SimReg, with approximately 12.5% average improvements for DDI predictions. To further investigate the difference among various methods, we provide a Student's t-test on the DTI and DDI results, respectively. Here, we assume that there is significant difference between two methods when the $p$-value is below 0.05. In Table S1, we summarize the $p$-value among single task-driven self-supervised representation learning (SSL) models for DTI and DTI predictions. We find that there is significant difference among most of methods. In particularly, all of the $p$-value between local information- and global information-based SSL models are below 0.05, as shown in the yellow shaded areas. Analogously, there is significant difference between attribute strong constraint- and weak constraint-based SSL models, as shown in the green shaded areas. These results further suggest that the global information (or attribute weak constraint)-driven SSL approaches significantly outperform the local information (or attribute strong constraint)-based SSL tasks.

## S2. Result analysis of two-task combinations in warm start scenarios

The results obtained by SSL on warm start drug predictions are shown in Table S2. Although PairDistance and SimCon generate great results, we find that PairDistance-SimCon shows the unsatisfactory performance (DDI-AUROC=0.880, and DTI-AUROC=0.942). In contrast, EdgeMask-PairDistance (DDI-AUROC=0.917, DTI-AUROC=0.958) and ClusterPre-PathClass (DDI-AUROC=0.915, DTI-AUROC= 0.956) produce relatively high results, with 2.0-5.9% higher AUROC and 2.8%-7.6% higher AUPR than other task combinations for DDI predictions. Concurrently, ClusterPre-PairDistance and EdgeMask-PathClass also produce promising results on DTI and DDI predictions. More interestingly, we find that EdgeMask-PairDistance, ClusterPre-PathClass, ClusterPre-PairDistance, and EdgeMask-PathClass are the combinations of global and local SSL tasks. In addition, PairDistance-SimCon (DDI-AUPR=0.860, DTI-AUPR=0.935) is superior to PairDistance-SimReg (DDI-AUPR= 0.847, DTI-AUPR=0.924). Similar situations are observed in comparison between SimCon and SimReg. Correspondingly, in Table S3, we summarize that the $p$-value among 11 two-modality combination models for DTI predictions. We observe that 9 out of 55 t-test experiments obtain $p$-value > 0.05. However, as shown in blue shaded area, there is significant difference among local-global combination models (i.e., EdgeMask-PairDistance, ClusterPre-PathClass, ClusterPre-PairDistance, and EdgeMask-PathClass) and other random combination models when there are the same sizes of modalities. We find a similar phenomenon in t-test experiments based on DDI results, as shown in Table S4. These results further suggest that the joint training of local and global SSL tasks tends to obtain higher performance than random two-task combinations when there are the same sizes of modalities.

## S3. Performance analysis of multi-task combinations in warm start scenarios

Although ClusterPre-PairDistance and EdgeMask-PathClass are local-global combination SSL models, as shown in Table S3 and Table S4, there is no significant difference ($p$-value>0.05) between them and other four combination models (i.e., PairDistance-PathClass, PathClass-SimCon, PairDistance-SimCon, and EdgeMask-SimReg). This may be attributed to the fact that EdgeMask-PathClass includes only single type of modality information (i.e., structures). However, PairDistance-PathClass and PathClass-SimCon capture two modalities of information (i.e., the structures and semantics of BioHNs). These results seem to indicate that we should consider the effect of different modalities. Therefore, we further design four combination models of multimodal tasks that refer to the structures, semantics and attributes of BioHNs.

We find that the top two combination models are ClusterPre-PathClass-SimReg and PairDistance-EdgeMask-SimCon in Table S2. Interestingly, they include three modalities of information, i.e., structure, semantic and attribute knowledge. Although EdgeMask-PathClass and EdgeMask-PairDistance belong to the local-global task combinations, EdgeMask-PairDistance is superior to EdgeMask-PathClass, with DDI-AUROC and DDI-AUPR improvements of approximately 3.5% and 4.3%, respectively. Similar phenomena is observed in comparison between ClusterPre-PathClass (DDI-AUROC=0.915, DDI-AUPR=0.910) and ClusterPre-PairDistance (DDI-AUROC= 0.895, DDI-AUPR=0.882). In other words, the combinations of two-modality tasks (e.g., EdgeMask-PairDistance and ClusterPre-PathClass) generate better results than the combinations of single-modality tasks (e.g., EdgeMask-PathClass and ClusterPre-PairDistance). More interestingly, we notice that PairDistance-SimReg-SimCon has one more task than PairDistance-SimReg. However, its DTI prediction performance exhibits no significant improvement. In contrast, for DDI predictions, PairDistance-SimReg-SimCon leads to a slight reduction compared to PairDistance-SimReg. This may be because PairDistance-SimReg-SimCon, in which the three tasks have only two-modality views (i.e., structure and semantic information), fails to increase the number of modalities over that used by PairDistance-SimReg and generates some noise. Similarly, ClusterPre-PairDistance-PathClass and ClusterPre-PathClass exhibit the same trend and phenomena. To further investigate the difference among 10 models, we conduct Student's t-test on the mixed results of DDI and DTI predictions. As shown in Table S5, there is significant difference ($p$-value <0.05) across various methods which include different modality information. The multi-task SSL models with multimodal information (e.g., PairDistance-EdgeMask-SimCon and ClusterPre-PathClass-SimReg) achieve greater results than other combination models. These results further suggest that combinations of multimodal tasks can achieve best performance for drug discovery.

## S4. Performance verification of MSSL2drug in cold start scenarios

For cold start scenarios, the results of single-task SSL models are shown in Fig. S1. PairDistance and PathClass yield better results than ClusterPre and EdgeMask. In particular, PathClass outperforms EdgeMask with 8.4% and 11.2% improvements in terms of AUROC and AUPR for DDI predictions, respectively. These results are straightforward and effective demonstrations that global information-based SSL can achieve better performance than local information-based SSL. Similarly, SimCon is superior to SimReg further suggesting that the attribute weak constraint-based SSL models outperform the strong constraint-based models.

In the two-task combination scenarios, as shown in Fig. S2 and Table S6, we observe the same phenomena as those exhibited in the warm start drug scenarios: the top three models are the local-global SSL tasks, that is, ClusterPre-PathClass (DDI-AUROC=0.890, DTI-AUROC=0.923), EdgeMask-PairDistance (DDI-AUROC=0.889, DTI-AUROC=0.927), and ClusterPre-PairDistance (DDI-AUROC=0.871, DTI-AUROC=0.911). These results further certify that the local and global SSL tasks jointly guide GNNs to generate superior drug discovery predictions when there are the same sizes of modalities.

In addition, the results of the multimodal tasks are shown in Fig. S3. We find that the SSL task combinations containing three modalities of information, such as ClusterPre-PathClass-SimReg (DDI-AUROC=0.909, DTI-AUROC=0.948) and PairDistance-EdgeMask-SimCon (DDI-AUROC=0.909, DTI-AUROC=0.940), are superior to the task combinations capturing two-modality knowledge, such as ClusterPre-PairDistance-PathClass (DDI-AUROC=0.894, DTI-AUROC=0.924), and PairDistance-SimReg-SimCon (DDI-AUROC=0.863, DTI-AUROC=0.918). Concurrently, the two-modality SSL task combinations outperform the one-modality SSL task combinations. In other words, as the number of modalities increases, the performance of cold start predictions is improved. These results further verify the multimodal combination strategy, that is, combinations of multimodal SSL tasks can achieve state-of-the-art the prediction performance of drug discovery.

## S5. Basic self-supervised tasks on BioHNs

**S5.1 Clustering coefficient regression (ClusterPre)**

In this work, we firstly design ClusterPre to develop the self-supervised representation learning (SSL) for drug discovery. In this pretext task, we aim to predict the clustering coefficient of each node to capture the local structure information in BioHNs. Formally, we adopt the mean squared error as the loss function of ClusterPre:

$$L_c(\theta) = \frac{1}{n}\sum_{i=1}^{n}\left(\delta\left(f_\theta(i)\right) - Y_{c_i}\right)^2 \tag{1}$$

where $\theta$ is the parameter of a graph neural network model $f_\theta(\cdot)$, $n$ represents the number of nodes, $f_\theta(i)$ denotes the representations of node $i$, $\delta(\cdot)$ is a Sigmoid function, and $Y_{c_i}$, which is the clustering coefficient for a given node $i$, can be calculated as follows:

$$Y_{c_i} = \frac{2l_i}{deg_i(deg_i - 1)} \tag{2}$$

where $deg_i$ is the degree of node $i$, and $l_i$ is the number of links between the $deg_i$ neighbors of node (i.e., the number of triangles that go through node $i$).

Generally, the clustering coefficients of nodes are larger when they have denser connections to other nodes. The closeness centrality can reflect the local structures in BioHNs to a large extent. The goal of ClusterPre is to ultimately learn the low-dimension representations that preserve the local structure information in BioHNs.

**S5.2 Pairwise distance classifications (PairDistance)**
The PairDistance self-supervised task is not limited to a node itself and its local neighborhoods; it also takes a global view of BioHNs. Three key steps as follows form the PairDistance task.

*Step1:* Randomly select a certain number of node pairs $(i, j)$ for which there is a path between nodes $i$ and $j$, and calculate the shortest path length $d_{i,j}$ for each node pair. This is mainly because calculating the shortest path lengths of all node pairs would be computationally expensive, and might be full of challenges for large-scale networks.

*Step2:* Divide all path lengths $d_{i,j}$ into four categories, that is, $d_{i,j} = 1, d_{i,j} = 2, d_{i,j} = 3$ and $d_{i,j} \geq 4$. Formally, we let $Y_{d_{i,j}} = \{d_{i,j} \mid d_{i,j} = 1, 2, 3, \text{and } d_{i,j} \geq 4\}$ denote the distance categories of node pairs.

*Step3:* Utilize GATs to predict the distance category of each node pair.

As described in *Step3*, PairDistance can be treated as a multiclass classification problem in which the objective function is formulated as follows:

$$L_{CD}(\theta) = -\frac{1}{|S|} \sum_{(i,j) \in S} \ell\left(\sigma\left(\langle f_\theta(i), f_\theta(j) \rangle\right), Y_{d_{i,j}}\right) \tag{3}$$

where $\langle \cdot, \cdot \rangle$ is an operation that concatenates two vectors, $\ell(\cdot, \cdot)$ represents the cross entropy loss function, and $\sigma(\cdot)$ represents the Softmax function. $S$ and $|S|$ denote the selected set of node pairs $(i, j)$ and the number of node pairs $(i, j)$, respectively.

**S5.3 Edge type masked predictions (EdgeMask)**
In this task, the edge representations, which are obtained by concatenating the representations of its two end-nodes, are fed into the Softmax function to predict the type of the masked edges. EdgeMask can be treated as the four classification problems. Similar to PairDistance, we also adopt the cross entropy loss function in EdgeMask. In the construed BioHN, there are five types of edges (e.g., drug-drug interactions, drug-protein interactions, drug-disease associations, protein-protein interactions, and protein-disease associations). However, previous studies [1-2] have suggested the schema of BioHNs, where both drug-drug relationships and protein-protein relationships are treated as 'interaction'. Concurrently, drug-drug networks and protein-protein networks are homogeneous networks. Inspired by these works, in MSSL2drug, drug-drug and protein-protein relationships are treated as the same types of semantic.

**S5.4 Bio-meta path classifications (PathClass)**
For a given heterogeneous network, a meta path is defined as a sequence in the form of $A_1 \xrightarrow{R_1} A_2 \xrightarrow{R_2} \cdots \xrightarrow{R_l} A_{l+1}$, which describes a composite relations between $A_1$ and $A_{l+1}$, where $A_l$ is node types, and $R_l$ represents edge type between nodes. In this work, the length of a meta path is defined as the total number of nodes in current meta path. A meta path integrates the semantic relationships in BioHNs. For example, "protein $\to$ disease $\to$ drug" describes the situation in which a protein causes a disease that is treated by a drug. Given a meta path, we can sample many path instances that have the same semantics, and belong to the same path type. Inspired by the multi-hub characteristics [1, 3] within BioHNs, we design 16 types of meta paths as shown in Table S8, where the first and last objects are drugs and proteins, respectively. This is mainly because drugs and proteins are interconnected with other entities by more edges, as shown in Table S9. Note that all meta paths include only four nodes because meta paths longer than four nodes may reduce the quality of associated semantic meanings.

In addition, we generate false path instances by randomly replacing some nodes in true path instances. There are three key points in the false path generation, as shown in Fig. S6, (1) For a given true path instance, it has 6.25% (i.e., 1/16) chance being generated a false path instance to avoid the label imbalance questions. (2) There is no relationship between the permutation nodes and the context nodes in current paths. (3) The number of replaced nodes is less than four that is the length of meta paths. In other words, there are at most two true edges in a false path. Therefore, these false paths can improve the generalization and robustness of PathClass when compared to the false paths with two true edges or one true edge. Nevertheless, it is interesting and important to verify what happens if the false paths have two true edges or one true edge.

**S5.5 Node similarity regression (SimReg)**
During the GNNs learning process, we perform node message propagation by aggregating local neighborhoods. Concurrently, we also wish to somewhat maintain the similarity attributes in the original feature space. Therefore, we develop SimReg, which requires GNNs to fit the similarity values of node pairs in the original feature space. Formally, the objective function employs the mean squared error and is given as follows:

$$L_{simReg}(\theta) = -\frac{1}{|S|} \sum_{(i,j) \in S} \ell'\left(\delta\left(\langle f_\theta(i), f_\theta(j) \rangle\right), Y_{sim_{i,j}}\right) \qquad (4)$$

where $\theta$ is the parameters of a graph neural network $f_\theta(\cdot)$, $f_\theta(i)$ and $f_\theta(j)$ denote the embeddings of node $i$ and $j$. $\langle \cdot, \cdot \rangle$ is an operation that concatenates two nodes, $\ell'(\cdot, \cdot)$ represents the mean square error (MSE) loss function, and $\delta(\cdot)$ represents the Sigmoid function. $Y_{sim_{i,j}}$ is the similarity value between two nodes in the original feature space. $S$ and $|S|$ denote the selected set of node pairs $(i, j)$ and the t5otal number of node pairs $(i, j)$, respectively.

In this work, we use different property similarity measurements according to various types of nodes. **Chemical similarities among drug pairs:** The simplified molecular input line entry system (SMILES) of each drug is extracted from DrugBank. For a given drug, we transform its SMILES sequence into an MACCS fingerprint by using Open Babel v2.3.1 (http://openbabel.org/wiki/Main_Page). Based on these MACCS fingerprints, we calculate the Tanimoto coefficient [4] of each drug-drug pair as its chemical similarity score. The Tanimoto coefficient offers a value in the range of zero to one and is widely used for drug discovery.

**Protein sequence similarity:** We download the protein sequences from the Uniprot database (http://www.uniprot.org/). We leverage the Smith-Waterman model [5] to calculate the sequence similarity scores of protein pairs. The Smith-Waterman algorithm performs local sequence alignment by comparing segments of all possible lengths and optimizing the similarity measure for determining similar regions between two strings of protein sequences.

**Disease similarity based on protein-protein interaction (PPI) networks:** The disease module theory [6] suggests that diseases with overlapping modules in gene-gene networks show significant symptom similarity and comorbidity. We calculate the disease similarity scores by using the ModuleSim algorithm [7-8], which is an extension of disease module theory.

$$sim(dis_1, dis_2) = \frac{2 * SIM(G_1, G_2)}{SIM(G_1, G_1) + SIM(G_2, G_2)} \qquad (5)$$

where $G_1 = \{g_{11}, g_{12}, \ldots, g_{1m}\}$ denotes a disease module, which contains $m$ genes for disease $dis_1$. $G_2$ is another disease module with a similar definition. $SIM(G_1, G_2)$ is calculated as follows:

$$SIM(G_1, G_2) = \frac{\sum_{1 \leq z \leq m_1} F_{G_2}(g_{1z}) + \sum_{1 \leq r \leq m_1} F_{G_1}(g_{2r})}{m_1 + m_2} \quad (6)$$

where $F_{G_2}(g_{1z}) = \text{avg}\left(\sum_{g \in G_2} sp(g_{1z}, g)\right)$. $sp(g_{1z}, g)$ is calculated as follows:

$$sp(g_{1z}, g) = \begin{cases} 1, & \text{if } g_{1z} = g \\ e^{-d_{g_{1z},g}}, & \text{otherwise} \end{cases} \quad (7)$$

where $d_{g_{1z},g}$ is the length of the shortest path between $g_{1z}$ and $g$ in a PPI network. $F_{G_1}(g_{2r})$ is also calculated according to similar definitions.

## S6. Hyperparameter setting

In SSL stage, we empirically consider the selections of optimization algorithm, weight initialization, and activation functions. The number of hidden layers, hidden units, head attentions, and batch size is selected according to the limitations of hardware (NVIDIA Tesla V100 GPU, Memory16G). The selections L2 regularization, learning rate, and epoch size are slightly tuning according to the performance of single tasks. Finally, we adopt the Glorot initialization [9], the Adam optimizer [10] with a learning rate [1e-5,1e-2], L2 regularization 5e-4, 8 hidden units and 8 head attentions. The number of epoch is set to 30. In supervised drug discovery, an MLP with three fully connected layers (including an input layer, a hidden layer and an output layer) is used to decode the embedding vectors. The size of the input layer depends on the dimensionality of the input feature, and the size of the hidden layer is set to 64. We also use the Adam optimizer to train the MLP for 30 epochs with batch size 128. For the learning rate, we select 10 points that are equidistant from the interval [5e-4, 5e-1].

  The representation dimension can directly affect the performance and time efficiency of self-supervised representation learning approaches. Therefore, several researchers have investigated the influence of different embedding dimensions in various SSL methods [11-15]. These studies found that the performance first increases when the embedding dimensionality increases. However, the performance tends to saturate or reduce when the dimension reaches to a threshold that is often close to 100. This is intuitive since higher dimensionality can encode more useful information, while too large value may lead to over-fitting phenomenon and excessive time complexity. In addition, the algorithm models with too large-scale parameters cannot run due to the limitations of hardware implementation. Therefore, the representation dimension of each private or shared GAT models is set to 64 (i.e., 8 hidden units $\times$ 8 head attentions) that is largest values under our hardware.

# Supplementary Figures

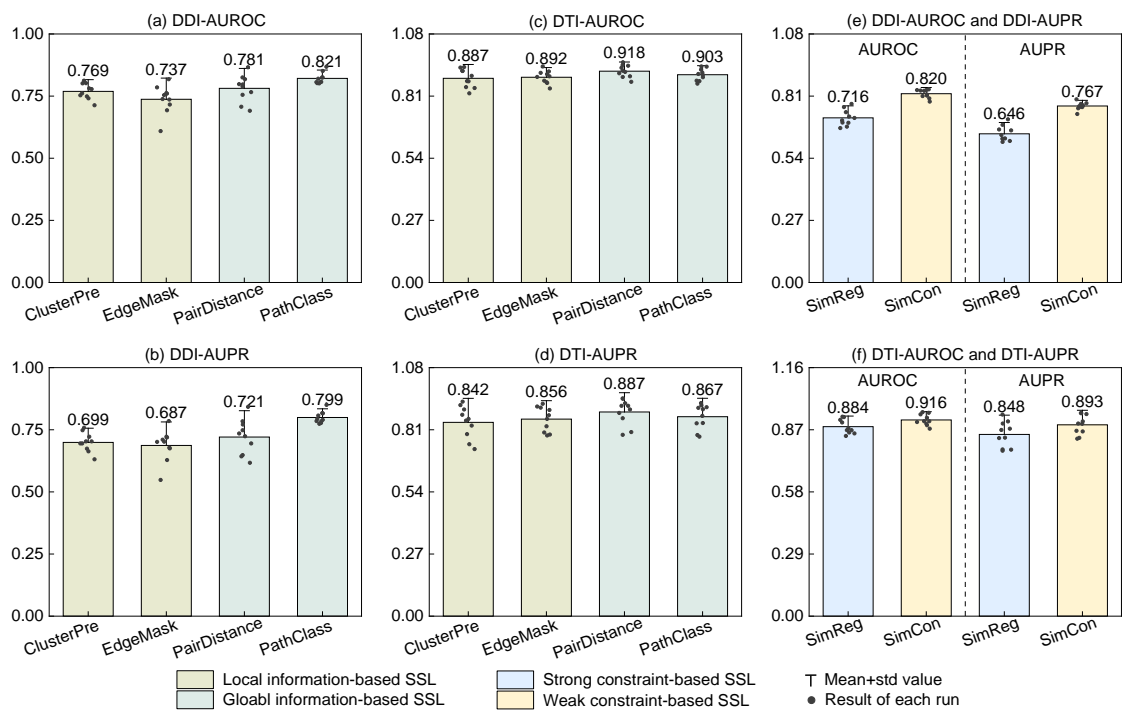

**Figure S1.** The results of single SSL tasks for cold start predictions where mean and std values denote average and standard deviation values that are calculated across ten results.

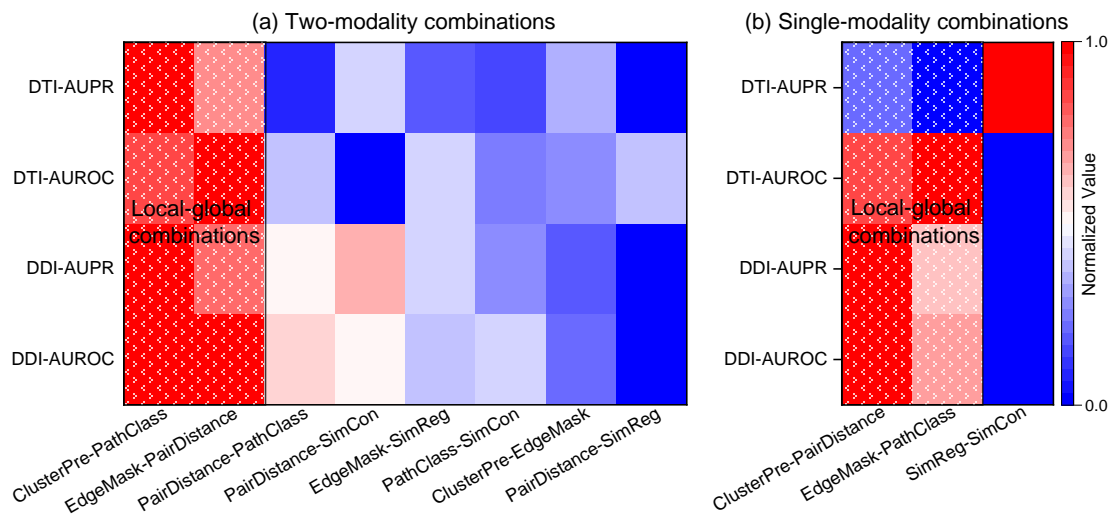

**Figure S2.** Heatmap of two-task combinations for cold start predictions where the results are normalized to [0,1] along the x-axis by Min-Max normalization technique. The redder (bluer) squares denote the greater (smaller) the value. The shaded area denotes the combinations of global and local SSL tasks.

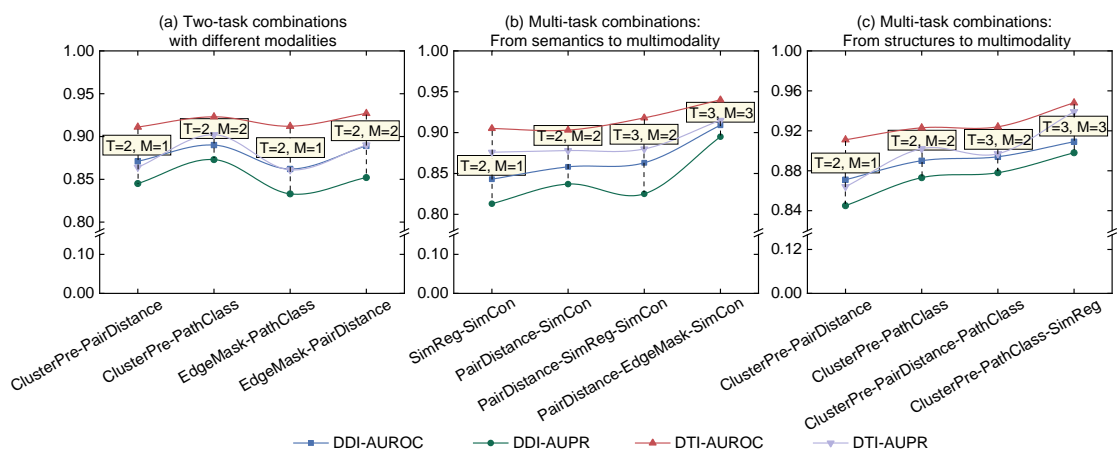

**Figure S3.** The results obtained with multimodal SSL tasks for cold start drug discovery, where 'T' and 'M' denote the total number of tasks and modalities in each multi-task combination, respectively.

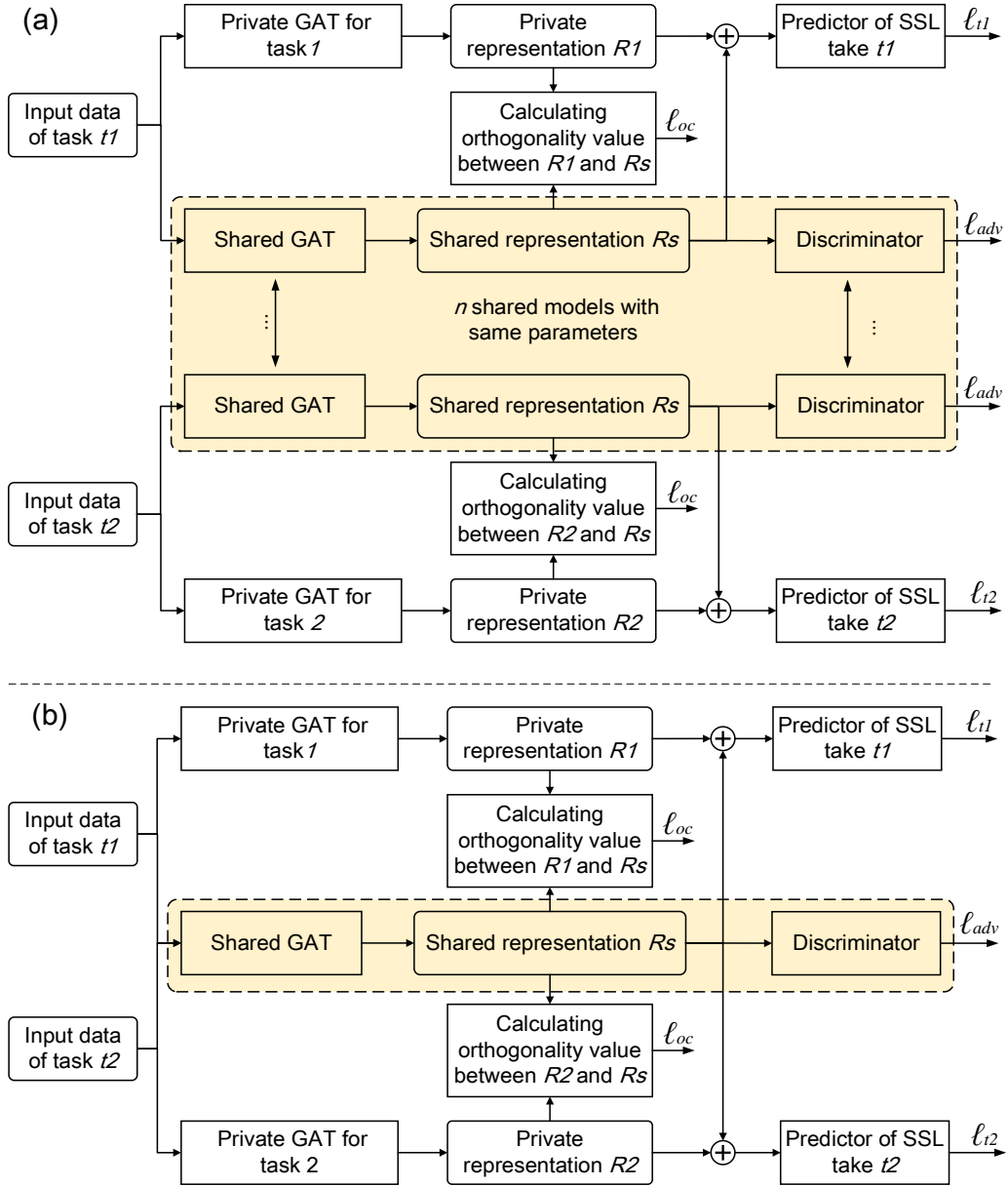

**Figure S4.** The frameworks of graph attention-based two-task learning.

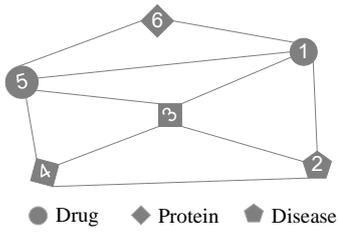

|        | Drug (X) | Protein (Y) | Disease (Z) | Degree    |
|--------|----------|-------------|-------------|-----------|
| Node 1 | 1        | 2           | 1           | 1+2+1=4   |
| Node 2 | 1        | 2           | 0           | 1+2+0=3   |
| Node 3 | 2        | 1           | 1           | 2+1+1=4   |
| Node 4 | 1        | 1           | 1           | 1+1+1=3   |
| Node 5 | 1        | 3           | 0           | 1+3+0=4   |
| Node 6 | 2        | 0           | 0           | 2+0+0=2   |

(a) Biomedical heterogeneous networks  
(b) Counting the number of neighbors in each class

One-hot encoding · One-hot encoding · One-hot encoding

|        | Drug matrix (X) | Protein matrix (Y) | Disease matrix (Z) |
|--------|-----------------|--------------------|--------------------|
| Node 1 | 1 0             | 0 0 1 0            | 0 1                |
| Node 2 | 1 0             | 0 0 1 0            | 1 0                |
| Node 3 | 0 1             | 0 1 0 0            | 0 1                |
| Node 4 | 1 0             | 0 1 0 0            | 0 1                |
| Node 5 | 1 0             | 0 0 0 1            | 1 0                |
| Node 6 | 0 1             | 1 0 0 0            | 1 0                |

|        | Initialization feature matrix (F) |
|--------|-----------------------------------|
| Node 1 | 1 0 0 0 1 0 0 1                   |
| Node 2 | 1 0 0 0 1 0 1 0                   |
| Node 3 | 0 1 0 1 0 0 0 1                   |
| Node 4 | 1 0 0 1 0 0 0 1                   |
| Node 5 | 1 0 0 0 0 1 1 0                   |
| Node 6 | 0 1 1 0 0 0 1 0                   |

(d) Initialization feature matrixes  
(c) One-hot encoding vector matrixes

**Figure S5.** An example of features initialization process.

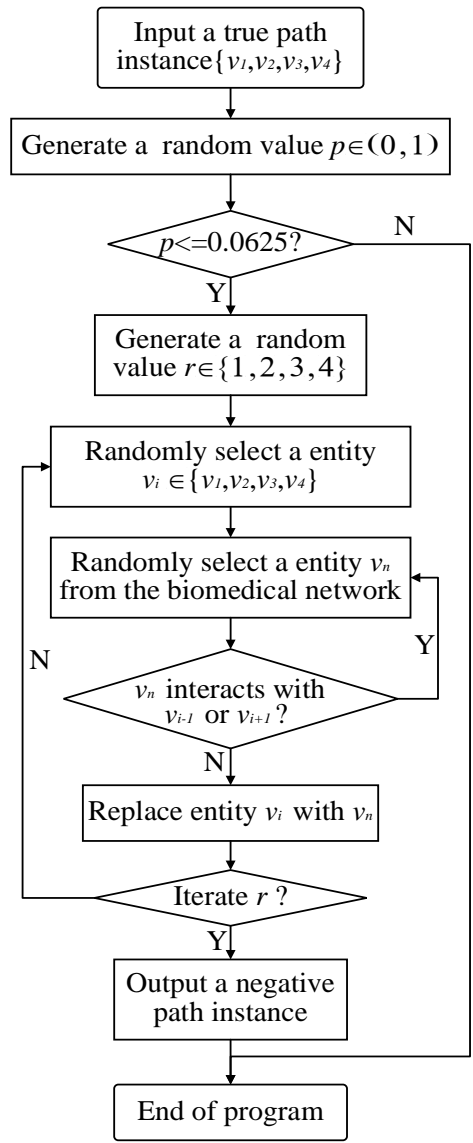

**Figure S6.** The procedure of negative path generation.

## Supplementary Tables

**Table S1.** The *p*-value among single task-driven SSL models for DTI and DDI predictions where we used the two-sided Student's t-test with a significance threshold of 0.05. No adjustments were made for multiple comparisons.

|     | Method | ClusterPre | EdgeMask | PairDistance | PathClass | SimReg | SimCon |
|-----|--------|------------|----------|--------------|-----------|--------|--------|
| DTI | ClusterPre | / | 0.112 | 2.38E-8 | 7.23E-5 | 5.41E-4 | 1.63E-5 |
|     | EdgeMask | 0.112 | / | 1.84E-3 | 4.66E-3 | 6.08E-7 | 3.30E-6 |
|     | PairDistance | 2.38E-8 | 1.84E-3 | / | 2.62E-3 | 0.815 | 5.17E-3 |
|     | PathClass | 7.23E-5 | 4.66 E-3 | 2.62E-3 | / | 1.59E-2 | 1.80E-5 |
|     | SimReg | 5.41E-4 | 6.08E-7 | 0.815 | 1.59E-2 | / | 3.58E-3 |
|     | SimCon | 1.63E-5 | 3.30E-6 | 5.17E-3 | 1.80E-5 | 3.58E-3 | / |
| DDI | ClusterPre | / | 1.50E-6 | 3.3E-4 | 2.68E-8 | 1.20E-5 | 1.54E-11 |
|     | EdgeMask | 1.50E-6 | / | 7.49E-9 | 3.50E-14 | 2.87E-12 | 3.75E-16 |
|     | PairDistance | 3.3E-4 | 7.49E-9 | / | 1.29E-6 | 0.121 | 9.35E-12 |
|     | PathClass | 2.68E-8 | 3.50E-14 | 1.29E-6 | / | 3.86E-8 | 2.84E-12 |
|     | SimReg | 1.20E-5 | 2.87E-12 | 0.121 | 3.86E-8 | / | 1.35E-15 |
|     | SimCon | 1.54E-11 | 3.75E-16 | 9.35E-12 | 2.84E-12 | 1.35E-15 | / |

a. The yellow shaded areas denote the *p*-value between local information- and global information-based SSL models.

b. The green shaded areas denote the *p*-value between attribute strong and weak constraint-based SSL models.

c. The *p*-value among SSL models higher than 0.05 are marked in red.

**Table S2.** The results obtained with fifteen SSL combinations in warm start predictions

| No. | Self-supervised tasks | Modal size | DDI | | DTI | |
|---|---|---|---|---|---|---|
| | | | AUROC±Std | AUPR±Std | AUROC±Std | AUPR±Std |
| 1 | EdgeMask-PairDistance | 2 | 0.917±0.003 | 0.912±0.005 | 0.958±0.007 | 0.951±0.009 |
| 2 | ClusterPre-PathClass | 2 | 0.915±**0.001** | 0.910±0.003 | 0.956±0.007 | 0.956±**0.007** |
| 3 | ClusterPre-PairDistance | 1 | 0.895±0.002 | 0.882±0.004 | 0.945±0.006 | 0.936±0.011 |
| 4 | EdgeMask-PathClass | 1 | 0.882±0.009 | 0.869±0.011 | 0.946±0.005 | 0.933±**0.007** |
| 5 | PairDistance-PathClass | 2 | 0.887±0.004 | 0.871±0.005 | 0.943±0.006 | 0.937±**0.007** |
| 6 | PathClass-SimCon | 2 | 0.883±0.004 | 0.867±0.006 | 0.947±0.009 | 0.945±0.010 |
| 7 | PairDistance-SimCon | 2 | 0.880±0.006 | 0.860±0.012 | 0.942±0.007 | 0.935±0.008 |
| 8 | EdgeMask-SimReg | 2 | 0.875±0.004 | 0.856±0.009 | 0.946±0.006 | 0.932±0.011 |
| 9 | PairDistance-SimReg | 2 | 0.873±0.004 | 0.847±0.010 | 0.936±**0.005** | 0.924±0.008 |
| 10 | ClusterPre-EdgeMask | 2 | 0.863±0.003 | 0.839±0.007 | 0.938±0.007 | 0.928±0.012 |
| 11 | SimReg-SimCon | 1 | 0.858±0.002 | 0.836±0.003 | 0.916±0.010 | 0.915±0.013 |
| 12 | ClusterPre-PairDistance-PathClass | 2 | 0.914±0.003 | 0.909±0.003 | 0.958±0.008 | 0.955±0.013 |
| 13 | ClusterPre-PathClass-SimReg | 3 | 0.926±0.004 | 0.923±0.005 | 0.968±0.016 | 0.966±0.018 |
| 14 | PairDistance-SimReg-SimCon | 2 | 0.879±0.008 | 0.858±0.016 | 0.944±0.007 | 0.938±0.010 |
| 15 | PairDistance-EdgeMask-SimCon | 3 | **0.939**±0.002 | **0.937**±**0.002** | **0.969**±0.006 | **0.968**±0.007 |

a. 'std' denotes the standard deviation value calculated across ten results.

b. The best results are marked in **boldface.**

**Table S3.** The *p*-value among 11 two-task combination models for DTI predictions where we used the two-sided Student's t-test with a significance threshold of 0.05. No adjustments were made for multiple comparisons.

| | EdgeMask-PairDistance | ClusterPre-PathClass | ClusterPre-PairDistance | EdgeMask-PathClass | PairDistance-PathClass | PathClass-SimCon | PairDistance-SimCon | EdgeMask-SimReg | PairDistance-SimReg | ClusterPre-EdgeMask | SimReg-SimCon |
|---|---|---|---|---|---|---|---|---|---|---|---|
| EdgeMask-PairDistance | / | 0.258 | 1.05 E-5 | 1.17 E-5 | 2.38 E-8 | 1.37 E-4 | 8.56 E-8 | 2.79 E-5 | 3.12 E-8 | 1.55 E-8 | 3.57 E-8 |
| ClusterPre-PathClass | 0.258 | / | 1.56 E-4 | 6.01 E-4 | 3.22 E-5 | 4.11 E-4 | 3.06 E-5 | 2.17 E-3 | 1.53 E-6 | 5.63 E-6 | 1.30 E-8 |
| ClusterPre-PairDistance | 1.05 E-5 | 1.56 E-4 | / | 0.935 | 0.025 | 0.582 | 0.029 | 0.661 | 7.51 E-5 | 3.43 E-4 | 2.39 E-6 |
| EdgeMask-PathClass | 1.17 E-5 | 6.01 E-4 | 0.935 | / | 0.015 | 0.700 | 0.011 | 0.466 | 8.62 E-6 | 5.87 E-4 | 5.25 E-6 |
| PairDistance-PathClass | 2.38 E-8 | 3.22 E-5 | 0.025 | 0.015 | / | 0.043 | 0.244 | 0.011 | 1.75 E-4 | 3.07 E-3 | 3.14 E-6 |
| PathClass-SimCon | 1.37 E-4 | 4.11 E-4 | 0.582 | 0.700 | 0.043 | / | 0.030 | 0.915 | 8.98 E-4 | 4.24 E-3 | 8.39 E-8 |
| PairDistance-SimCon | 8.56 E-8 | 3.06 E-5 | 0.029 | 0.011 | 0.244 | 0.030 | / | 4.41 E-3 | 1.20 E-3 | 0.013 | 4.74 E-6 |
| EdgeMask-SimReg | 2.79 E-5 | 2.17 E-3 | 0.661 | 0.466 | 0.011 | 0.915 | 4.41 E-3 | / | 3.66 E-5 | 1.97 E-4 | 6.05 E-6 |
| PairDistance-SimReg | 3.12 E-8 | 1.53 E-6 | 7.51 E-5 | 8.62 E-6 | 1.75 E-4 | 8.98 E-4 | 1.20 E-3 | 3.66 E-5 | / | 0.257 | 2.00 E-5 |
| ClusterPre-EdgeMask | 1.55 E-8 | 5.63 E-6 | 3.43 E-4 | 5.87 E-4 | 3.07 E-3 | 4.24 E-3 | 0.013 | 1.97 E-4 | 0.257 | / | 3.29 E-5 |
| SimReg-SimCon | 3.57 E-8 | 1.30 E-8 | 2.39 E-6 | 5.25 E-6 | 3.14 E-6 | 8.39 E-8 | 4.74 E-6 | 6.05 E-6 | 2.00 E-5 | 3.29 E-5 | / |

a. The blue shaded areas represent the *p*-value between local-global combination models and other models.
b. The *p*-value among SSL models higher than 0.05 are marked in red.

**Table S4.** The *p*-value among 11 two-task combination SSL models for DDI predictions where we used the two-sided Student's t-test with a significance threshold of 0.05. No adjustments were made for multiple comparisons.

| | EdgeMask-PairDistance | ClusterPre-PathClass | ClusterPre-PairDistance | EdgeMask-PathClass | PairDistance-PathClass | PathClass-SimCon | PairDistance-SimCon | EdgeMask-SimReg | PairDistance-SimReg | ClusterPre-EdgeMask | SimReg-SimCon |
|---|---|---|---|---|---|---|---|---|---|---|---|
| EdgeMask-PairDistance | / | 0.145 | 1.53 E-9 | 3.80 E-7 | 7.03 E-9 | 1.70 E-9 | 1.46 E-10 | 3.71 E-12 | 4.26 E-12 | 7.77 E-14 | 7.79 E-14 |
| ClusterPre-PathClass | 0.145 | / | 2.67 E-10 | 1.43 E-6 | 5.92 E-10 | 2.90 E-9 | 4.22 E-12 | 4.72 E-10 | 3.85 E-10 | 9.97 E-15 | 4.27 E-13 |
| ClusterPre-PairDistance | 1.53 E-9 | 2.67 E-10 | / | 2.03 E-3 | 3.55 E-4 | 5.96 E-05 | 5.33 E-8 | 1.90 E-7 | 9.92 E-8 | 6.40 E-12 | 5.18 E-12 |
| EdgeMask-PathClass | 3.80 E-7 | 1.43 E-6 | 2.03 E-3 | / | 0.304 | <span style="color:red">0.813</span> | <span style="color:red">0.418</span> | 0.025 | 0.011 | 5.59 E-5 | 5.86 E-6 |
| PairDistance-PathClass | 7.03 E-9 | 5.92 E-10 | 3.55 E-4 | 0.304 | / | <span style="color:red">0.056</span> | 4.12 E-3 | 2.87 E-5 | 7.13 E-6 | 2.53 E-8 | 8.86 E-9 |
| PathClass-SimCon | 1.70 E-9 | 2.90 E-9 | 5.96 E-5 | <span style="color:red">0.813</span> | <span style="color:red">0.056</span> | / | <span style="color:red">0.062</span> | 2.38 E-4 | 1.64 E-4 | 2.19 E-7 | 3.03 E-8 |
| PairDistance-SimCon | 1.46 E-10 | 4.22 E-12 | 5.33 E-8 | <span style="color:red">0.418</span> | 4.12 E-3 | <span style="color:red">0.062</span> | / | 9.02 E-3 | 2.03 E-3 | 9.54 E-9 | 2.57 E-9 |
| EdgeMask-SimReg | 3.71 E-12 | 4.72 E-10 | 1.90 E-7 | 0.025 | 2.87 E-5 | 2.38 E-4 | 9.02 E-3 | / | 0.014 | 4.19 E-6 | 4.33 E-8 |
| PairDistance-SimReg | 4.26 E-12 | 3.85 E-10 | 9.92 E-8 | 0.011 | 7.13 E-6 | 1.64 E-4 | 2.03 E-3 | 0.014 | / | 1.98 E-5 | 1.05 E-7 |
| ClusterPre-EdgeMask | 7.77 E-14 | 9.97 E-15 | 6.40 E-12 | 5.59 E-5 | 2.53 E-8 | 2.19 E-7 | 9.54 E-9 | 4.19 E-6 | 1.98 E-5 | / | 1.21 E-3 |
| SimReg-SimCon | 7.79 E-14 | 4.27 E-13 | 5.18 E-12 | 5.86 E-6 | 8.86 E-9 | 3.03 E-8 | 2.57 E-9 | 4.33 E-8 | 1.05 E-7 | 1.21 E-3 | / |

a. The blue shaded areas represent the *p*-value between local-global combination models and other models.

b. The *p*-value among SSL models higher than 0.05 are marked in red.

**Table S5.** The *p*-value among 10 SSL models based on the mixed results of DDI and DTI predictions where we used the two-sided Student's t-test with a significance threshold of 0.05. No adjustments were made for multiple comparisons.

| | ClusterPre-PathClass | EdgeMask-PairDistance | ClusterPre-PairDistance | EdgeMask-PathClass | PairDistance-SimCon | SimReg-SimCon | PairDistance-SimReg-SimCon | PairDistance-EdgeMask-SimCon | ClusterPre-PairDistance-PathClass | ClusterPre-PathClass-SimReg |
|---|---|---|---|---|---|---|---|---|---|---|
| ClusterPre-PathClass | / | 0.145 | 2.67 E-10 | 1.43 E-6 | 4.22 E-12 | 4.27 E-13 | 1.26 E-7 | 1.99 E-12 | 0.061 | 8.62 E-7 |
| EdgeMask-PairDistance | 0.145 | / | 1.53 E-9 | 3.80 E-7 | 1.46 E-10 | 7.79 E-14 | 4.94 E-8 | 2.42 E-9 | 0.017 | 5.87 E-5 |
| ClusterPre-PairDistance | 2.67 E-10 | 1.53 E-9 | / | 0.002 | 5.33 E-8 | 5.18 E-12 | 2.15 E-4 | 1.66 E-12 | 1.67 E-7 | 3.22 E-11 |
| EdgeMask-PathClass | 1.43 E-6 | 3.80 E-7 | 0.002 | / | 0.418 | 5.86 E-6 | 0.263 | 8.79 E-9 | 3.04 E-6 | 4.18 E-7 |
| PairDistance-SimCon | 4.22 E-12 | 1.46 E-10 | 5.33 E-8 | 0.418 | / | 2.57 E-9 | 0.451 | 2.63 E-14 | 1.39 E-10 | 1.20 E-10 |
| SimReg-SimCon | 4.27 E-13 | 7.79 E-14 | 5.18 E-12 | 5.86 E-6 | 2.57 E-9 | / | 7.45 E-6 | 4.34 E-15 | 2.33 E-12 | 1.28 E-12 |
| PairDistance-SimReg-SimCon | 1.26 E-7 | 4.94 E-8 | 2.15 E-4 | 0.263 | 0.451 | 7.45 E-6 | / | 1.00 E-9 | 5.33 E-8 | 1.19 E-7 |
| PairDistance-EdgeMask-SimCon | 1.99 E-12 | 2.42 E-9 | 1.66 E-12 | 8.79 E-9 | 2.63 E-14 | 4.34 E-15 | 1.00 E-9 | / | 2.16 E-10 | 7.00 E-7 |
| ClusterPre-PairDistance-PathClass | 0.061 | 0.017 | 1.67 E-7 | 3.04 E-6 | 1.39 E-10 | 2.33 E-12 | 5.33 E-8 | 2.16 E-10 | / | 1.55 E-5 |
| ClusterPre-PathClass-SimReg | 8.62 E-7 | 5.87 E-5 | 3.22 E-11 | 4.18 E-7 | 1.20 E-10 | 1.28 E-12 | 1.19 E-7 | 7.00 E-7 | 1.55 E-5 | / |

a. The gray shaded areas represent the *p*-value between multimodal combinations and other models.

b. The *p*-value among SSL models higher than 0.05 are marked in red.

**Table S6.** The results obtained with fifteen SSL combinations in cold start predictions

| No. | Self-supervised tasks | Modal size | DDI | | DTI | |
|---|---|---|---|---|---|---|
| | | | AUROC±Std | AUPR±Std | AUROC±Std | AUPR±Std |
| 1 | ClusterPre-PathClass | 2 | 0.890±0.017 | 0.873±0.024 | 0.923±0.028 | 0.902±0.054 |
| 2 | EdgeMask-PairDistance | 2 | 0.889±0.014 | 0.852±0.017 | 0.927±0.030 | 0.890±0.077 |
| 3 | ClusterPre-PairDistance | 1 | 0.871±0.018 | 0.845±0.026 | 0.911±0.027 | 0.864±0.074 |
| 4 | EdgeMask-PathClass | 1 | 0.862±0.018 | 0.833±0.025 | 0.912±0.024 | 0.861±0.057 |
| 5 | PairDistance-PathClass | 2 | 0.862±0.021 | 0.825±0.028 | 0.912±0.028 | 0.864±0.067 |
| 6 | PairDistance-SimCon | 2 | 0.858±0.016 | 0.837±0.020 | 0.903±0.037 | 0.878±0.065 |
| 7 | EdgeMask-SimReg | 2 | 0.848±0.022 | 0.814±0.026 | 0.913±0.029 | 0.867±0.072 |
| 8 | PathClass-SimCon | 2 | 0.850±0.020 | 0.801±0.035 | 0.909±0.032 | 0.866±0.079 |
| 9 | ClusterPre-EdgeMask | 2 | 0.837±0.021 | 0.792±0.029 | 0.910±0.026 | 0.875±0.058 |
| 10 | PairDistance-SimReg | 2 | 0.822±0.054 | 0.774±0.070 | 0.912±0.029 | 0.860±0.073 |
| 11 | SimReg-SimCon | 1 | 0.843±0.019 | 0.813±0.026 | 0.905±0.028 | 0.876±0.085 |
| 12 | ClusterPre-PairDistance-PathClass | 2 | 0.894±0.017 | 0.878±0.027 | 0.924±0.027 | 0.897±0.061 |
| 13 | ClusterPre-PathClass-SimReg | 3 | **0.909**±0.013 | **0.898**±0.016 | **0.948**±0.022 | **0.939**±0.052 |
| 14 | PairDistance-SimReg-SimCon | 2 | 0.863±0.021 | 0.825±0.024 | 0.918±0.024 | 0.880±0.060 |
| 15 | PairDistance-EdgeMask-SimCon | 3 | **0.909**±**0.008** | 0.895±**0.011** | 0.940±**0.020** | 0.915±**0.048** |

The best results are marked in **boldface**.

**Table S7.** The numbers of nodes and edges in the constructed BioHN

| Type of node | Count | Type of edge | Count |
|---|---|---|---|
| Drug | 721 | Drug-Drug interactions | 66,384 |
| Protein | 1,894 | Drug-protein interactions | 4,978 |
| Disease | 431 | Drug-disease associations | 1,201 |
| / | / | Protein-protein interactions | 16,133 |
| / | / | Disease-protein associations | 23,080 |
| Total | 3046 | Total | 111,776 |

**Table S8.** The types of meta paths

| NO. | Meta path |
|---|---|
| 1 | drug-drug-drug-protein |
| 2 | drug-drug-protein-protein |
| 3 | drug-drug-disease-protein |
| 4 | drug-protein-drug-protein |
| 5 | drug-protein-protein-protein |
| 6 | drug-protein-disease-protein |
| 7 | drug-disease-drug-protein |
| 8 | drug-disease-protein-protein |
| 9 | protein-drug-drug-drug |
| 10 | protein-protein-drug-drug |
| 11 | protein-disease-drug-drug |
| 12 | protein-drug-protein-drug |
| 13 | protein-protein-protein-drug |
| 14 | protein-disease-protein-drug |
| 15 | protein-drug-disease-drug |
| 16 | protein-protein-disease-drug |

**Table S9.** The total number of edges connected to drugs, proteins, and diseases, respectively.

| Node types | Drugs | Proteins | Diseases |
| --- | --- | --- | --- |
| The total number of edges | 72,563 | 44,191 | 24,281 |

**Table S10.** The examples of multi-task combinations with different modalities

| Modal size | Multi-task combinations with different modalities | |
| --- | --- | --- |
| 1 | SimReg-SimCon | ClusterPre-PairDistance |
| 2 | PairDistance-SimCon | ClusterPre-PathClass |
| 2 | PairDistance-SimReg-SimCon | ClusterPre-PairDistance-PathClass |
| 3 | PairDistance-EdgeMask-SimCon | ClusterPre-PathClass-SimReg |